\documentclass[10pt,twocolumn,letterpaper]{article}

\usepackage[final]{pdfpages}
\usepackage{titlesec}
\usepackage{cvpr}
\usepackage{times}
\usepackage{epsfig}
\usepackage{graphicx}
\usepackage{amsmath}
\usepackage{amssymb}
\usepackage{graphicx}
\usepackage{epstopdf}
\usepackage{subfig}
\usepackage{makecell}
\usepackage{multirow}
\usepackage{array}
\usepackage{bm}
\usepackage{dsfont}
\usepackage{color}
\usepackage{caption}
\usepackage{algorithm} 
\usepackage{algorithmic} 
\usepackage{xcolor}

\usepackage{pifont}
\newcommand{\cmark}{\ding{51}}%
\newcommand{\xmark}{\ding{55}}%

\usepackage[pagebackref=true,breaklinks=true,letterpaper=true,colorlinks,bookmarks=false]{hyperref}

\cvprfinalcopy 

\def\cvprPaperID{0009} 

\ifcvprfinal\fi
\begin{document}

\title{Exploiting Offset-guided Network for Pose Estimation and Tracking}

\author{Rui Zhang$^{*1,2}$, Zheng Zhu$^{*3}$, Peng Li$^{1}$, Rui Wu$^{4}$, Chaoxu Guo$^{*3}$, Guan Huang$^{4}$, Hailun Xia$^{*1,2}$\\
$^{1}$ Bejing Key Laboratory of Network System Architecture and Convergence,\\ School of Information and Communication Engineering, \\Beijing University of Posts and Telecommunications, Beijing.\\
$^{2}$ Beijing Laboratory of Advanced Information Networks, Beijing.\\
$^{3}$ Institute of Automation, Chinese Academy of Sciences, Beijing.\\
$^{4}$ Horizon Robotics, Beijing.\\
{\tt\small zhengzhu@ieee.org}
}

\maketitle
\thispagestyle{empty}

\begin{abstract}
Human pose estimation has witnessed a significant advance thanks to the development of deep learning. Recent human pose estimation approaches tend to directly predict the location heatmaps, which causes quantization errors and inevitably deteriorates the performance within the reduced network output. Aim at solving it, we revisit the heatmap-offset aggregation method and propose the Offset-guided Network (OGN) with an intuitive but effective fusion strategy for both two-stages pose estimation and Mask R-CNN. For two-stages pose estimation, a greedy box generation strategy is also proposed to keep more necessary candidates while performing person detection. For mask R-CNN, ratio-consistent is adopted to improve the generalization ability of the network. State-of-the-art results on COCO and PoseTrack dataset verify the effectiveness of our offset-guided pose estimation and tracking.
\end{abstract}

\footnotetext{*The first two authors contributed equally to this work.}

\section{Introduction}
Human pose estimation in images and articulated pose tracking in videos are of significance for visual understanding task \cite{zhou2014learning, MaskRCNN,zhu2018endto,zhu2018two}. Research community has witnessed a significant advance from single person \cite{PS,DPM,DeepPose,tompson2014joint,CPM,Hourglass,fppose} to multi-person pose estimation \cite{DeepCut,DeeperCut,OpenPose,GooglePose,CPN}, from static images pose estimation \cite{pishchulin2012articulated, MaskRCNN} to articulated tracking in videos \cite{Posetrack,ArtTrack,JointFlow,kristan2018sixth,bai2018multi,SiamRPN,DaSiamRPN,Detect_and_track,PoseFlow,MSRAPose}. However, there are still challenging pose estimation problems in complex environments, such as occlusion, intense light and rare poses \cite{MPII,COCO,diagnosispose}. Furthermore, articulated tracking encounters new challenges in unconstrained videos such as camera motion, blur and view variants \cite{PoseTrackBenchmark, UCT, FlowTrack}.

Previous pose estimation systems address single pre-located person, which exploit pictorial structures model\cite {PS,DPM} and deep convolutional neural network \cite{DeepPose,tompson2014joint,CPM,Hourglass,fppose}. Motivated by practical applications in video surveillance, human-computer interaction and action recognition, researchers now focus on the multi-person pose estimation in unconstrained environments. Multi-person pose estimation can be categorized into bottom-up \cite{DeepCut,DeeperCut,OpenPose} and top-down approaches \cite{GooglePose,CPN,MaskRCNN,MSRAPose}, where the latter becomes dominant participants in COCO benchmarks \cite{COCO}. Top-down approaches can be divided into two-stages based methods and unified framework. Two-stages methods \cite{GooglePose,CPN,MSRAPose} firstly detect and crop persons
from the image, then perform the single person pose estimation in the cropped person patches. Representative work of unified framework methods is Mask R-CNN \cite{MaskRCNN}, which extracts the human bounding box and predicts keypoints from the corresponding feature maps simultaneously.

While there has been a significant advance in pose estimation, quantization errors still exist in most of the modern networks. Although Google \cite{GooglePose} proposes to simultaneously classify the heatmaps and regress the offset filed, recent human pose estimation approaches \cite{MaskRCNN,CPN,MSRAPose} tend to directly predict the location heatmaps. Because of the quantization effect between input and output, performance is inevitably deteriorated within the reduced network output. While both deconv and offset can reduce quantization errors, offset is more significant for resources-restricted applications due to its efficiency. In this paper, we revisit the heatmap-offset aggregation method and propose the Offset-guided Network (OGN) for both two-stages pose estimation and unified Mask R-CNN framework. We extend modern frameworks by adding a branch for offset prediction in parallel with the existing branch. Meanwhile, an intuitive but effective fusion is adopted to obtain the final results, and we propose a greedy box generation strategy to keep more necessary candidates. The OGN aims at improving precision for all sizes output especially low resolution. Our network can output keypoints location in continuous space which reduces the quantization error.

In experiments, the offset-guided two-stages pose estimation approach reaches mAP of 74.0 on COCO test-dev set, yielding 14\% relative gain compared with \cite{GooglePose}. On PoseTrack dataset, we achieve 67.7 MOTA using two-stages pose input without optical flow, which is the new state-of-the-art results in this task.

The main contributions can be described as follows:

 (1) Heatmap-offset aggregation method is revisited and we propose the OGN for both two-stages pose estimation and Mask R-CNN. An intuitive but effective fusion strategy is adopted to obtain the final results by merging two branches.

 (2) As a novel alternative to NMS, a greedy box generation strategy is adopted to keep more necessary candidates for offset-guided two-stages pose estimation.

 (3) In experiments, the offset-guided two-stages pose estimation
approach reaches mAP of 74.0 on COCO test-dev set with a single model, yielding 14\% relative gain compared to \cite{GooglePose}. Furthermore,
we achieve 67.7 MOTA on PoseTrack dataset without optical flow, which is the new state-of-the-art results in this task.

\section{Related Works}

\subsection{Single person pose estimation}
Single person pose estimation is a task that predicts the pose of a single person in an image. Conventional methods \cite{PS,DPM} exploit pictorial structure model which expresses the human body as a tree-structured graphical model. \cite{PS} claims that the right selection of components for both appearance and spatial modeling is crucial. The Deformable Part Model (DPM) \cite{DPM} adopts HOG feature to implement this idea. Recently, this task has been advanced rapidly for the development of deep convolution neural networks. \cite{DeepPose} firstly tries to utilize CNN and they prefer to directly regress coordinates of body parts. More recently, researches on this task choose to regress some heat maps, which each stands for a body part. \cite{tompson2014joint} is the first work which solves the problem by using CNN and graphical models to predict heat maps of each body part. With the continuous work of many researchers, some novel architectures like CPM \cite{CPM}, Stacked Hourglass \cite{Hourglass} and PRMs \cite{fppose} are used to achieve state-of-the-art results.

\subsection{Multi-person pose estimation}
Motivated by practical applications, researchers now focus on multi-person in unconstrained environments. Multi-person pose estimation can be categorized into bottom-up and top-down approaches where the latter becomes dominant participants in COCO benchmarks \cite{COCO}.

\paragraph{bottom-up}Bottom-up architecture based methods first detect body parts and then associate corresponding body parts with specific human instances. The typical methods are DeepCut \cite{DeepCut} and DeeperCut \cite{DeeperCut}. The former adopts an integer linear programming based method and the later improves DeepCut via utilizing image-conditioned pairwise terms. \cite{OpenPose} predicts heatmaps of body parts and a set of 2D vector fields of part affinities and parses them by greedy inference to generate the final results.

\paragraph{Top-down}Top-down approaches can be divided into two-stages based methods and unified framework. Two-stages methods \cite{pishchulin2012articulated,GooglePose,CPN,MSRAPose} first detect and crop persons from an image, then perform single person pose estimation in the cropped person patches. \cite{pishchulin2012articulated} follows this two-step framework by using pictorial structure models based method. \cite{GooglePose} combines classification and regression tasks which respectively predicts the offset vector and location heatmap of each body part. \cite{CPN} proposes a cascaded pyramid network containing global pyramid network and pyramid refined network which aims for online hard key points mining. Representative work of unified framework methods is Mask R-CNN \cite{MaskRCNN} that builds an end-to-end framework and yields an impressive performance.

\begin{figure*}[ht]
\centering
\includegraphics[scale=0.45]{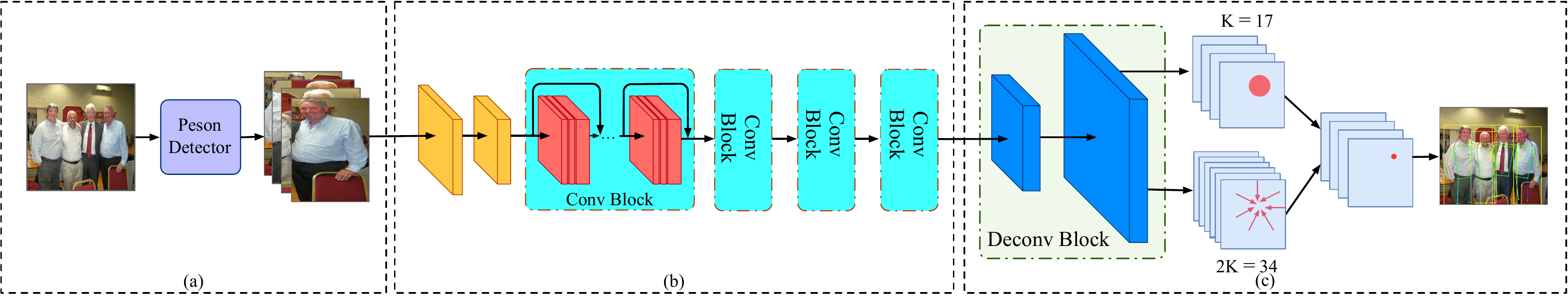}
\caption{Offset-guided two-stages pose estimation network. It consists of three main components: (a) the person detector, (b) extracting features using ResNet, (c) the process of refinement and fusion}
\label{fig:OGN}
\end{figure*}

\section {Overview of offset-guided network}

For pose estimation, it is noticed that the precision of keypoints localization is limited by the size of network output. During the downsampling process, there exists a quantization error. OGN is utilized to address this problem. We verify the effectiveness of OGN for two-stages pose estimation (shown in Figure \ref{fig:OGN}) and extended Mask R-CNN framework (Figure \ref{fig:MaskRcnn}). Following \cite{GooglePose}, the regions of interest (ROIs) detected and cropped by person detector are fed to the pose estimator, where the offset regression branch guides heatmap classification branch to refine the pose location. Differently, two \textit{deconv} layers \cite{MSRAPose} are used to enlarge the heatmaps by four times and an intuitive but effective fusion is adopted to obtain the final results. Meanwhile, in extended Mask R-CNN as shown in Figure \ref{fig:MaskRcnn}, the ROIs from RPN are firstly extended to a fixed ratio and then ROI-Align is used to extract the feature in each extended ROIs. Finally, a score map and an offset map are predicted and fused to obtain the final location of keypoints.

\subsection{Offset-guided two-stages Pose Estimation}\label{two-stages}
 We first address the OGN for two-stages pose estimation framework. For the first stage, the results of the person detector are crucial for subsequent pose estimator. However, the box with a lower score may have higher IoU with ground truth and may be eliminated by the subsequent NMS \cite{NMS} process. Therefore, a Greedy Box Generation (GBG) strategy is proposed to retain more necessary candidates. For the second stage, two branches are used to obtain the final results. The offset regression branch guides heatmap classification branch to approach the ground truth. Meanwhile, the heatmap classification branch guides offset regression branch to focus on the neighborhood of ground truth.


\subsubsection{Greedy Box Generation Strategy}
We adopt Mask R-CNN \cite{MaskRCNN} as the person detector that achieves AP 51.7 of 80 categories detection on the COCO val2017. Different from most of the other approaches, we propose a greedy box generation (GBG) strategy as a novel alternative to Non-Maximum Suppression (NMS) \cite{NMS}. It prefers to retain redundant boxes which helps us to get better pose selected by OKS+IOU NMS after pose estimation. Specifically, no filtering strategies including NMS are used in both RPN and R-CNN phase. As a result of person detection, thousands of boxes are put out as candidates. The sequential selection of candidates can be described as follows. Firstly, based on the task limitation, we filter out the boxes whose size are smaller than the minimum threshold. Then, those boxes whose confidence score is larger than 0.8 are picked out. We argue those boxes are reliable and call them equivalent ground truth (EGT). After that, the other predicted boxes who has a $IoU<0.5$ with all EGT are eliminated. Finally, all of the rest boxes are divided into groups where every box has a $IoU>=0.7$ with each other, and top $N$ of each group (we use $N = 4$) are preserved. By adopting GBG strategy, we tend to keep the boxes with score relatively small but localization more accurate.

\subsubsection{Offset-guided Network}
In this work, we utilize ResNet \cite{ResNet} as the backbone of the offset-guided network. Our offset-guided architecture addresses two main problems. Firstly, in order to preserve more local details, \textit{deconv} layers are appended for higher resolution. In our practice, two \textit{deconv} layers are used to enlarge the feature maps by four times. Secondly, following \cite{GooglePose}, we adopt an approach combining classification and regression branches to obtain the final pose results which helps to reduce quantization errors. We denote the number of keypoints by $K$. A convolution layer of $K=17$ channels is adopted to output coarse location, and a convolution layer of $2K$ channels to regress the offset for a fine position. For each predicted position $x_i$ and each GT key point $g_k$, the target label for the classification head is:
\begin{equation}
H_c = \left\{\begin{matrix}
 1 &  ||x_i-g_k||<= R & \\
 0 &  ||x_i-g_k||> R&
\end{matrix}\right.
\end{equation}
The target label for the $x$-axis of offset is:
\begin{equation}
H_r = \left\{\begin{matrix}
(g_k-x_i)/R &  ||x_i-g_k||<= R & \\
 0 &  ||x_i-g_k||> R&
\end{matrix}\right.
\end{equation}
And the same is $y$-axis. The classification head considers the whole heatmaps, while the offset loss is only computed within a disk of radius $R$ from each keypoint. Our insight is that these two heads can revise each other. The regression head helps to revise the coarse location of keypoints. The classification head helps to exclude the invalid regions, so the regression head can focus on learning offset within a small range. Besides, this heatmap-offset aggregation method outputs result in continuous space which eliminates the quantization errors. As shown in Figure \ref{fig:OGN}, the OGN can be split into three stages. In experiments, the OGN dramatically improves the performance in a large range of output resolutions, especially for low resolution.

\subsubsection{Inference}
\paragraph{}Inspired by \cite{fang2017rmpe}, to make the pose estimator adapted to the boxes generated by our person detector, we mix up the predicted boxes and ground truth boxes. With this strategy, our pose estimator can adapt to the variance of box location distribution and perform better while testing. Once those ROIs are provided, the cropped areas from the original image will be sent to a single pose estimator. In our practice, ResNet is used to extract features and some \textit{deconv} layers \cite{MSRAPose} is added to pursue higher resolution. Smooth $L_1$ is used as the loss function for both classification and regression. In addition, we employ a Gaussian filter to make the output heatmaps smoother. The final results are obtained by merging two branches using an intuitive but effective fusion method.
\paragraph{}For classification branch, each key point is predicted by a heatmap $L_k \in R_{W*H}$ ($W,H$ is the width and height of the final heatmap respectively). The other branch is used to generate $2K$ heatmaps. Every pair of them stands for the $x, y$ offset for the corresponding position in $L_k$ , and these heatmaps are denoted by $O_{kx}, O_{ky}$. Each $O_{kx}$ and $O_{ky}$ have the same size with $L_k$. Firstly, we find the maximum score in each $L_k$ and mark them as coarse localization $(T_w, T_h)$.
{\setlength\abovedisplayskip{4pt}
\setlength\belowdisplayskip{4pt}
\begin{equation}
\label{eq1}
\textbf {$(T_w, T_h)$} = argmax(\textbf {$L_k$}), k=1,2,...,k
\end{equation}}
Then, the corresponding offsets $O_{{kx}_w}$, $O_{{ky}_h}$ in $O_{kx}$ and $O_{ky}$ are obtained. Finally, the output can be denoted as:
{\setlength\abovedisplayskip{4pt}
\setlength\belowdisplayskip{4pt}
\begin{equation}
\label{eq1}
(\textbf {$F_x, F_y$}) = (\textbf {$T_w+O_{{kx}_w}R, T_h+O_{{ky}_h}R$})
\end{equation}}

\vspace{-0.5cm}
\begin{figure}[ht]
\centering
\includegraphics[height=0.25\textheight, width=0.5\textwidth]{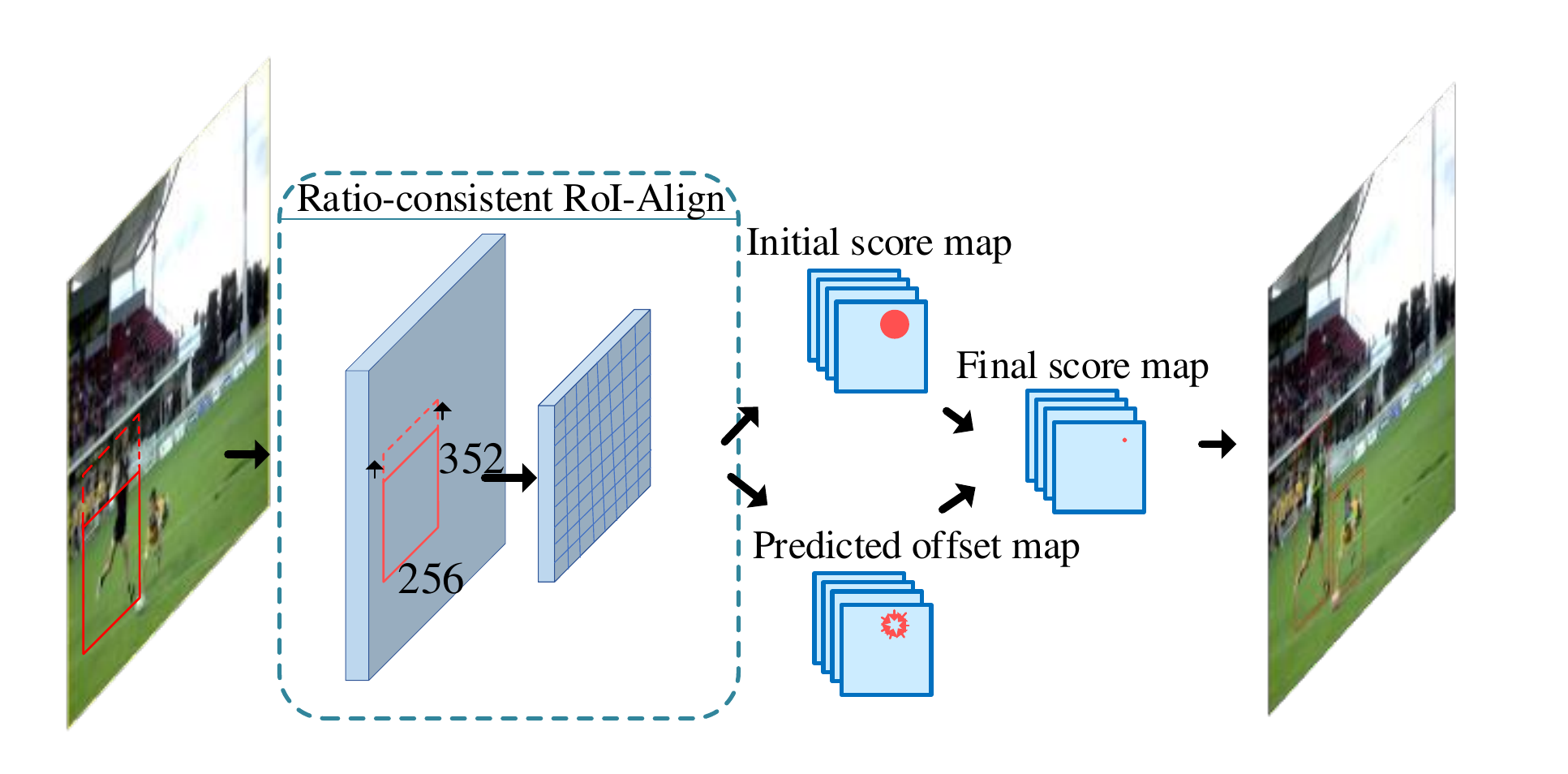}
\caption{The framework of the offset-guided Mask R-CNN. The ROIs from RPN are firstly extended to a fixed ratio and then ROI-Align is used to extract the feature in each extended ROIs. Then a score branch and an offset branch are predicted and fused to obtain the final location of keypoints.}

\label{fig:MaskRcnn}
\end{figure}

\vspace{-0.8cm}
\subsubsection{Discussion}
 Compared to \cite{GooglePose}, our method emphasizes simplicity and effectiveness. \cite{GooglePose} adopts logistic loss for the classification head and Hober robust loss for the regression head, while only the Smooth $L_1$ loss is used for both of them in this paper. The totally different loss types in \cite{GooglePose} introduce a hyper-parameter to keep loss balanced. In contrast, we only need to simply add the loss of the two branches together. When it comes to the process of fusion, \cite{GooglePose} adopts Hough voting strategy while we directly select the maximum prediction. What is more, our network can still converge well without any intermediate supervision while \cite{GooglePose} adds an extra heatmap to contribute auxiliary loss. From these perspectives, OGN can be easily transferred to other frameworks like Mask R-CNN \cite{MaskRCNN}. Our approach is not only simple and intuitive, but also effective. In Section ~\ref{sec:pose exp}, comprehensive experiments are conducted to verify the effectiveness.

\subsection{Offset-guided Mask R-CNN}
Besides the above two-stages pose estimation, the effectiveness of the OGN is also evaluated on Mask R-CNN, which is an end-to-end framework producing results of detection and pose estimation simultaneously. \cite{MaskRCNN} models the location of a keypoint as a one-hot mask and produces $K$ masks for each keypoint based on the feature from ROI-Align. However, ROI-Align will output distorted feature map in different degrees if the ratios of ROIs are different, which increases the training difficulty of subsequent prediction head. And the resulting one-hot map may be less accurate due to the small resolution of the feature map.

Therefore, this paper proposes two techniques to improve the performance of human pose estimation of Mask R-CNN. The first one is to transform all human ROIs into a fixed ratio by extending the width or height of ROIs, which makes sure the ROIs fed into prediction head fall into the same distribution of ratio and improve the ability of generalization. This strategy is denoted as \textit{ratio-consistent} in the following sections. The second one is that the human pose is predicted with a score map and an offset map. Specially, the prediction with the max response in the score map represents the coarse prediction location, and the offset map further refines it to a finer location. Here the score map is the same as the score map in single person pose estimation model mentioned in Section~\ref{two-stages} and we also use Smooth $L_1$ loss as the optimization target. The extended Mask R-CNN framework is illustrated in Figure~\ref{fig:MaskRcnn}.

\section{Experiments}
\begin{table*}[!tp]
\setlength{\tabcolsep}{5pt}
\scriptsize
  \centering
  \captionsetup{font={normalsize}}
  \caption{Ablation study on the COCO \textit{val2017} set (* indicates that flip test is used). Method (\textit{a}) reduces the number of \textit{deconv} layers to one in MSRA \cite{MSRAPose} published source code. Method (\textit{b}) comes from \cite{MSRAPose}. Based on our network, method (\textit{c}-\textit{l}) conduct experiments on offset, GBG strategy, input size and  \textit{deconv} layers.}
\vspace{-0.3cm}
\begin{tabular}{ccccccccccccc}\label{tab:ablation COCO val2017}
\\\hline
\bf Method & \bf Network & \bf \makecell[cc]{Input \\ Size}  &\bf \textit{Deconv}& \bf  \makecell[cc]{Feature \\ Stride} & \bf Offset  &\bf GBG & \boldmath $AP$ & \boldmath $AP_{0.5}$ & \boldmath $AP_{0.75}$ & \boldmath $AP_m$ &\boldmath $AP_l$ \\
\hline
\textit{a} & ResNet-50(MSRA)* & 256x192 & 1 & 16 & \xmark & \xmark & 52.0 & 88.8 & 58.2 & 51.5 & 53.1 \\\hline
\textit{b} & ResNet-50(MSRA)* & 256x192 & 2 & 8 & \xmark & \xmark & 68.2 & - & - & - & - \\\hline
\textit{c} & ResNet-50 & 256x192 & 2 & 8 & \cmark & \xmark & 69.7 & 88.2 & 77.2 & 66.0 & 75.8 \\\hline
\textit{d} & ResNet-50 & 256x192 & 1 & 16 & \cmark& \xmark & 67.7 & 90.3 & 74.6 & 64.5& 72.9  \\\hline
\textit{e} & ResNet-50 & 352x256 & 1 & 16 & \cmark& \cmark & 70.4 & 90.5 & 76.4 & 67.0 & 75.8 \\\hline
\textit{f} & ResNet-50 & 384x288 & 1 & 16 & \cmark& \cmark & 70.7 & 88.5 & 77.4 & 66.3 & 77.6 \\\hline
\textit{g} & ResNet-50 & 512x384 & 1 & 16 & \cmark& \cmark & 71.7 & 88.7 & 77.7 & 67.0 & 79.0 \\\hline
\textit{h} & ResNet-50 & 512x384 & 1 & 16 & \cmark& \xmark & 71.3 & 88.6 & 77.7 & 67.3 & 78.5 \\\hline
\textit{i} & ResNet-50 & 384x288 & 2 & 8 & \cmark& \cmark & 71.6 & 89.0 & 78.3 & 67.3 & 78.5 \\\hline
\textit{j} & ResNet-50 & 512x384 & 2 & 8 & \cmark& \cmark & 73.0 & 91.5 & 79.6 & 68.8 & 78.8 \\\hline
\textit{k} & ResNet-101 & 512x384 & 2 & 8 & \cmark& \cmark & 73.8 & 91.6 & 79.6 & 70.0 & 79.7 \\\hline
\textit{l} & ResNet-152 & 512x384 & 2 & 8 & \cmark& \cmark & 74.0 & 91.5 & 79.7 & 70.1 & 79.9
\\\hline
\end{tabular}
   \label{Coco-val2017}
\end{table*}

In this section, the performance of the proposed offset-guided network is evaluated on COCO and PoseTrack dataset.

\subsection{Results on COCO dataset}\label{sec:pose exp}
\begin{figure*}[tp]
\centering
\includegraphics[height=0.47\textheight, width=1\textwidth]{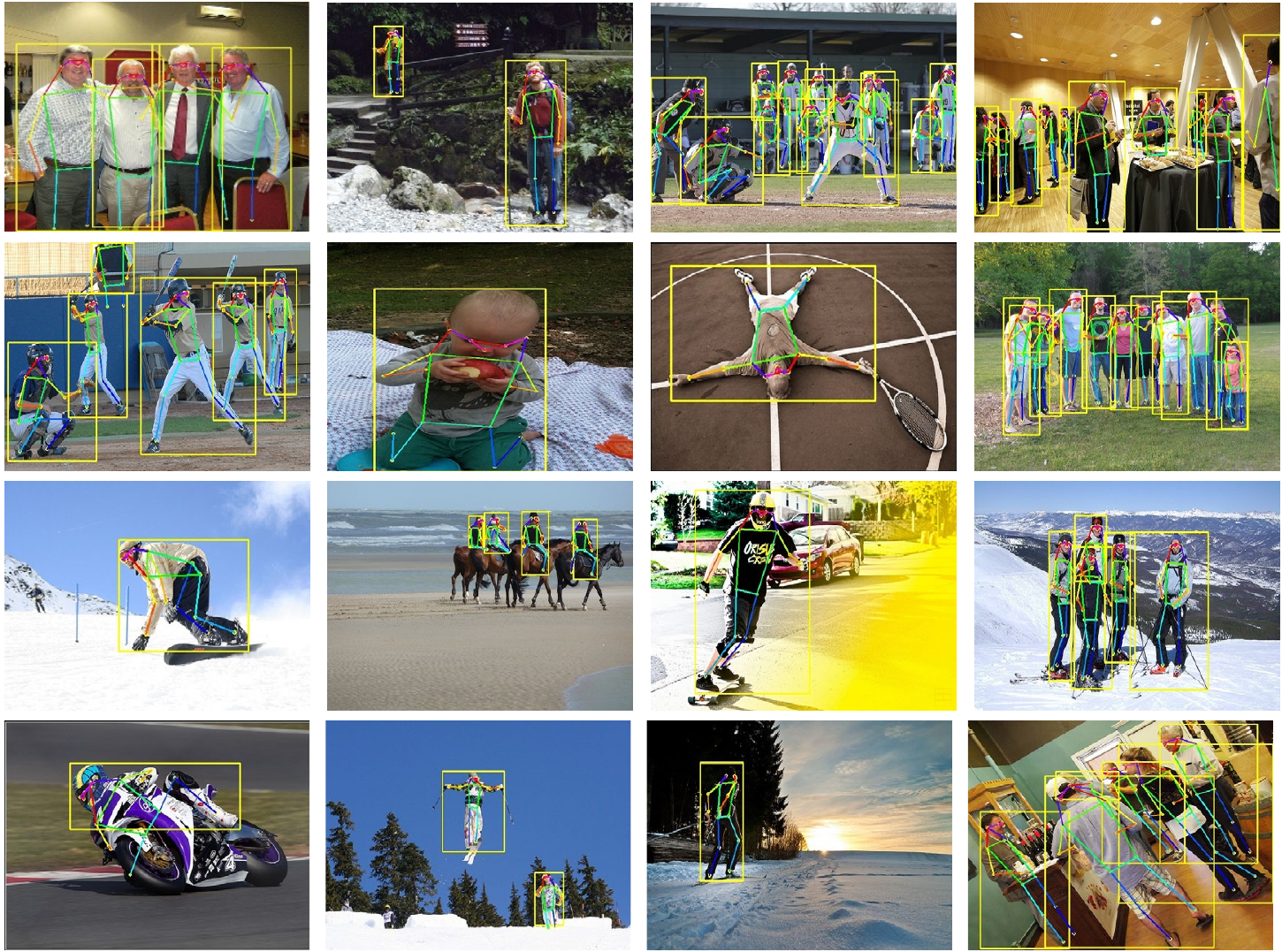}
\caption{Qualitative results of the COCO dataset. In each frame, bounding box, keypoints and skeleton are denoted by rectangle, dot and line separately.
}
\label{fig:coco}
\end{figure*}

Experiments are firstly conducted on the COCO \cite{COCO} benchmark which requires both person detection and body parts localization in uncontrolled conditions. The COCO dataset contains more than 200k images and 250k person instances splitting into train, validation and test sets. Ablation study is conducted on the validation set. To compare with other methods, we provide final results on both test-dev and test-challenge2018.  The qualitative results of the COCO dataset are shown in Figure \ref{fig:coco}.
\paragraph{Train details}
Our offset-guided two-stages model is pre-trained on the Imagenet \cite{Imagenet} classification dataset. For data augmentation, random flip, rotation ($\pm30^\circ$) and scale (0.9$\sim$1.2) on original image are adopted. Considering the peculiarity of multi-person pose estimation task, we use a ROI based sampling strategy to improve the model's generalization ability. Eight TITAN X GPUs and batch size of 64 are used. For every iteration, we randomly choose two images for each GPU and four ROIs for each image. The whole train process contains 22 epochs. The learning rate is 0.02 and drops twice at the 17th epoch and the 21st epoch with the decay of 0.1, SGD optimizer is used.

\paragraph{Test details}
The test is conducted on the COCO val2017, test-dev and test-challenge2018. Following our GBG strategy, all ROIs generated by detected boxes are adjusted to a fixed ratio 3:4. For post-processing, a Gaussian filter is used to smooth the heatmaps at first. Then following \cite{CPN}, we use the product of box score and pose score as the final score for the sorting mechanism. Finally, NMS \cite{NMS} based on $IoU=0.6$ and $OKS=0.75$ is employed.

\begin{figure*}[ht]
\centering
\includegraphics[height=0.6\textheight, width=1\textwidth]{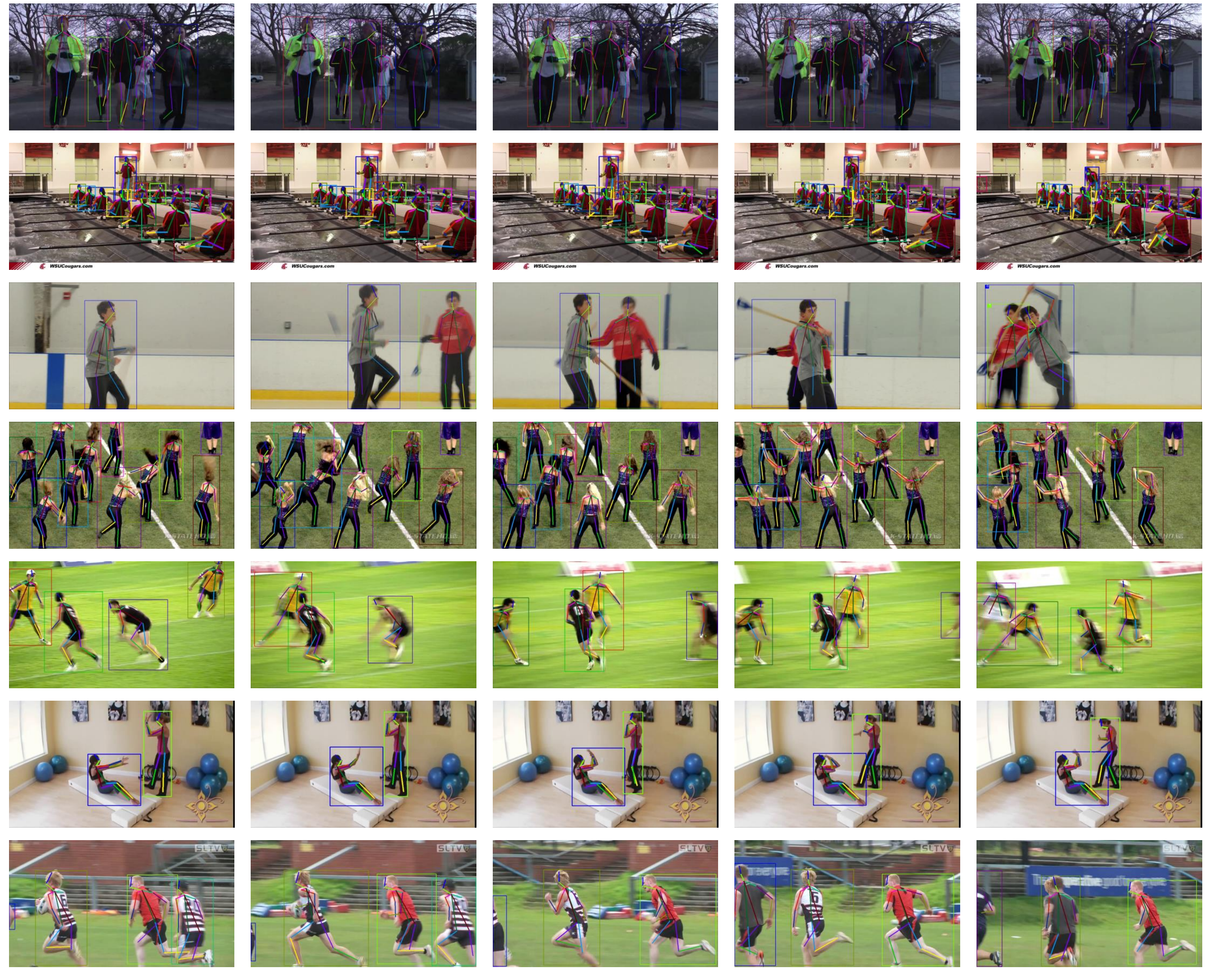}
\caption{
Qualitative results of the PoseTrack dataset. In each frame, bounding box and pose of the human are illustrated, where the same color boxes indicate the same identity.
}
\label{fig:flow}
\end{figure*}

\subsubsection{Ablation study}
Ablation study is conducted on the COCO val2017 set. Offset, GBG, Resolution and network depth are considered as shown in Table \ref{tab:ablation COCO val2017}.
\begin{enumerate}
    \item \textbf{Offset} Network with low resolution output is of significance for resources-restricted applications due to its efficiency. From method (\textit{a}, \textit{d}), it can be seen that $stride=16$ will inevitably deteriorate the performance if offset is not considered. Performance can be improved by 15.7 AP when considering offset. When $stide=8$, our OGN method (\textit{c}) improves MSRA baseline method (\textit{b}) by 1.5 AP. As shown in Table \ref{tab:coco-dev-maskrcnn}, our offset-guided architecture also improves Mask R-CNN by 2.1 AP.
    \item \textbf{GBG} From the comparison of methods (\textit{g}, \textit{h}), the AP can be improved by 0.4 using our GBG strategy.
    \item \textbf{Resolution} Resolution is affected by input size and network stride. Comparing methods (\textit{e}, \textit{f}, \textit{g}) with each other, one can find that larger network input produces better results within certain range. As input size grows, the AP increases by 0.3 and 1.0 respectively. \textit{Deconv} layers can reduce the network stride as shown in methods (\textit{f}, \textit{i}) and (\textit{g}, \textit{j}). Similarly with \cite{MSRAPose}, AP increases by 0.9 from method \textit{f} to \textit{i}. When adopting larger input size, our final AP can increase by 1.3 from method \textit{g} to \textit{j}.
    \item \textbf{Network depth} Comparison of methods (\textit{j}, \textit{k}, \textit{l}) exposes that performance benefits from deeper network. Changing network depth, AP can increase by 0.8 from ResNet-50 to ResNet-101 and 1.0 from ResNet-50 to ResNet-152.
\end{enumerate}

\subsubsection{Comparison with state-of-the-art results}
The proposed OGN method participates in both COCO Keypoints 2017 and 2018 challenges. In 2017, the performance of our single model is 71.3 AP, and our final result reaches 72.8 AP on \textit{test-dev} set when state-of-the-art is 73.0. In 2018, as shown in Table \ref{tab:coco-dev}, our single model method, without additional training data and ensemble, achieves new state-of-the-art performance on the COCO \textit{test-dev} set with 74.0 AP, which yields 14\% relative gain compared to \cite{GooglePose}. Comparing with the previous state-of-the-arts \cite{MSRAPose}, our approach improves the results by 0.2 AP. Our result with 100k additional data and the ensemble of ResNet \cite{ResNet}, ResNext \cite{Liu2017Skeleton}, Xception \cite{Chollet2016Xception} achieves 75.9 AP. In \textit{test-challenge} set, our result ranks the \textit{3rd} by 74.1 AP among COCO leaderboard when submitted.

\begin{table}[!ht]
\setlength{\tabcolsep}{1.8pt}
\scriptsize
  \centering
  \captionsetup{font={normalsize}}
  \caption{Pose estimation performance with single model on the COCO \textit{test-dev} set.}
   \vspace{-0.35cm}
\begin{tabular}{ccccccc}\label{tab:coco-dev}
\\\hline
\bf Method &  \boldmath $AP$ & \boldmath $AP_{0.5}$ & \boldmath $AP_{0.75}$ & \boldmath $AP_m$ &\boldmath $AP_l$ & \bf AR \\
\hline
\textbf {CMU-Pose}\cite{OpenPose} & 61.8 & 84.9 & 67.5 & 57.1 & 68.2 & - \\\hline
\textbf {Mask-RCNN}\cite{MaskRCNN} & 63.1 & 87.3 & 68.7 & 57.8 & 71.4 & - \\\hline
\textbf {G-RMI}\cite{GooglePose} & 64.9  & 85.5 & 71.3 & 62.3 & 70.0 & 69.7 \\\hline
\textbf {CPN}\cite{CPN} & 72.1 & 90.5 & 78.9 & 67.9 & 78.1 & 78.7 \\\hline
\textbf {MSRA}\cite{MSRAPose} & 73.8 & 91.7 & 81.2 & 70.3 & 80.0 & 79.1 \\\hline
\textbf {Ours-2017} & 71.3 & 91.0 & 78.3 & 67.9 & 76.3 & 74.4 \\\hline
\textbf {Ours-2018} & 74.0 & 91.1 & 81.1 & 69.8 & 80.5 & 79.7 \\\hline
\end{tabular}
   \label{coco-test-dev}
   \vspace{-0.2cm}
\end{table}

\begin{table}[!ht]
\setlength{\tabcolsep}{1.8pt}
\scriptsize
  \centering
  \captionsetup{font={normalsize}}
  \caption{Compared with Mask R-CNN on the \textit{test-dev} set.}
  \vspace{-0.3cm}
\begin{tabular}{ccccccc}\label{tab:coco-dev-maskrcnn}
\\\hline
\bf Method &  \boldmath $AP$ & \boldmath $AP_{0.5}$ & \boldmath $AP_{0.75}$ & \boldmath $AP_m$ &\boldmath $AP_l$ & \bf AR \\\hline

\textbf {Mask R-CNN}\cite{MaskRCNN} & 63.1 & 87.3 & 68.7 & 57.8 & 71.4 & - \\\hline
\textbf {Extended Mask R-CNN} & 65.2 & 88.2 & 70.9 & 61.8 & 71.7 & 69.8 \\\hline
\end{tabular}
   \label{coco-maskrcnn}
\end{table}

\subsection{Results on PoseTrack dataset}

Experiments about pose estimation and tracking on PoseTrack dataset are also conducted. Ablation study on offset-guided Mask R-CNN is conducted. Similar tracking strategy as \cite{Detect_and_track} is adopted except that the appearance information is taken into account. Specifically, we utilize the metric which integrates the spatial cue and the appearance cue. IoU is adopted to measure the spatial similarity and human Re-identification model is utilized to extract the appearance feature of the targets. Furthermore, the Euclidean distance is adopted to measure the appearance similarity.

\subsubsection{Ablation study}
We evaluate the proposed ratio-consistent strategy and offset-guided Mask R-CNN on PoseTrack \textit{val2017} dataset. The results are illustrated in Table \ref{Mask R-CNN PoseTrack val2017}. We conduct experiments in three aspects including offset, ratio-consistent and the type of loss to optimize.
\begin{enumerate}
   \item \textbf{Offset} Comparing method a with \cite{Detect_and_track}, the performance improvement is 3.7 mAP, which proves the effectiveness of OGN.
    \item \textbf{Ratio-consistent} Similar to single person pose, the ratio of ROIs in this model is extended to $352\times256$. It brings another 1.6 mAP improvement comparing to method a.
    \item \textbf{Loss} We regress the score maps of keypoint location and offset map with Smooth L1 loss. With this technique, 66.7 mAP is obtained. Our final results on val2017 outperform \cite{Detect_and_track} by 0.8 mAP.
\end{enumerate}

\begin{table}[!th]
\scriptsize
  \centering
  \captionsetup{font={normalsize}}
  \caption{Ablation study of extended Mask R-CNN on PoseTrack \textit{val2017} set. The backbone is ResNet-101. The result of FAIR is from \cite{Detect_and_track}.}
\begin{tabular}{ccccc}\label{Mask R-CNN PoseTrack val2017}
\\\hline
  \bf \makecell[cc] {Method}&
  \bf \makecell[cc] {Loss Type}&\bf \makecell[cc]{Ratio\\Consistent}  & \boldmath \bf \makecell[cc]{Offset-guided \\Refinement}  & \bf \boldmath \makecell[cc]{mAP \\ Total}  \\
\hline
 \textbf{FAIR}  & softmax  & \xmark  & \xmark & 60.6 \\
 \textit{a} & softmax  & \xmark  & \checkmark & 64.3 \\
 \textit{b} & softmax  & \checkmark  & \checkmark & 65.9 \\
 \textit{c} & Smooth $L_1$      & \checkmark  & \checkmark & 66.7  \\
\hline
\end{tabular}
   \label{Coco-val2017}
\end{table}

\subsubsection{Comparison with state-of-the-art results}
As shown in Table \ref{pose track performance}, without optical flow, there is an improvement of MOTA over existing best method \cite{MSRAPose} by 2.3 on PoseTrack \textit{val2017}. If the optical flow is adopted, the MOTA improvement is 4.7. Meanwhile, Our approach obtains state-of-the-art performance on \textit{test2017} set. The qualitative results of the PoseTrack dataset are shown in Figure \ref{fig:flow}.

\begin{table}[!th]

\scriptsize
  \centering
  \captionsetup{font={normalsize}}
  \caption{Multi-person pose estimation and tracking performance on PoseTrack 2017 dataset. We adopt the same optical flow method as MSRA.}
  \vspace{-0.2cm}
\begin{tabular}{cccc} \label{pose track performance}
\\\hline
\bf \makecell[cc]{Method} & \bf \makecell[cc]{Dataset}  & \boldmath \makecell[cc]{Total \\ mAP}  & \boldmath \makecell[cc]{Total \\ MOTA}\\
\hline
\textbf{MSRA} \cite{MSRAPose} & \textit{validation}   & 76.7 & 65.4\\
\textbf{FAIR} \cite{Detect_and_track} & \textit{validation} & 64.1 &55.2 \\
\textbf{PoseFlow} \cite{PoseFlow} & \textit{validation} &66.5 &58.3\\
\textbf{Ours (Mask R-CNN)} & \textit{validation}     & 66.7 &60.7 \\
\textbf{Ours (two stages)} & \textit{validation}     &75.1 &67.7 \\
\textbf{Ours (two stages with optical flow)} & \textit{validation}     &76.7 &70.1 \\ \hline
\textbf{MSRA} \cite{MSRAPose} & \textit{test}     & 74.6 & 57.8 \\
\textbf{FAIR} \cite{Detect_and_track} & \textit{test}     &- & 51.8\\
\textbf{PoseFlow} \cite{PoseFlow} & \textit{test}  &63.0 &51.0\\
\textbf{Ours(Mask R-CNN)} & \textit{test} &63.9 &57.4   \\
\textbf{Ours(two stages)} & \textit{test} &72.6 &59.2  \\
\textbf{Ours(two stages with optical flow)} & \textit{test} &74.8 &61.6   \\ \hline
\end{tabular}
\label{table:pose_estimate}
\end{table}
\vspace{-0.3cm}

\section{Conclusion}
In this paper, we revisit the heatmap-offset aggregation method for pose estimation and propose the Offet-guided network (OGN) for both two-stages approaches and Mask R-CNN. The OGN is designed to reduce errors caused by the quantization effect between network input and output. A novel alternative to NMS for two-stages network is proposed which named GBG. For offset-guided Mask R-CNN, ratio-consistent is adopted to improve the model's ability of generalization. State-of-the-art results are achieved on both COCO and PoseTrack dataset.

{\small
\bibliographystyle{ieee}
\bibliography{egbib}
}
\end{document}